\documentclass[journal]{IEEEtran}
\usepackage{amsmath, amssymb, amsthm}
\usepackage{graphicx}
\usepackage{algorithmic}
\usepackage{caption}
\usepackage{subcaption}
\usepackage{cite}
\usepackage{booktabs}
\usepackage[scaled=1.0]{helvet}
\usepackage{times}
\usepackage{graphicx}

\usepackage{parskip}
\usepackage{multirow}
\usepackage{amsmath}
\usepackage{color}

\usepackage{booktabs}
\usepackage{multirow}
\usepackage{color}
\usepackage{amssymb}
\usepackage{balance}
\usepackage{bm}
\usepackage{cite}

\usepackage{graphicx}
\usepackage{caption}
\usepackage{subcaption}

\usepackage[table, xcdraw]{xcolor}
\usepackage{makecell}

\usepackage{threeparttable}

\usepackage[labelfont=bf,textfont=it]{caption}
\usepackage[lined,boxed,commentsnumbered, ruled]{algorithm2e}

\begin{document}

\title{Neural Block-wise Auto-exposure for Single-Shot High Dynamic Range Imaging}

\author{Zhengguo Li,~\IEEEmembership{Fellow,~IEEE},    Chaobing Zheng,~\IEEEmembership{Member~IEEE}, and Wei Wang,~\IEEEmembership{Member~IEEE}

\thanks{Zhengguo Li  is with Institute of Advanced Intelligence and Computing (IAIC), Agency for Science, Technology and Research (A*STAR), Singapore, 138632, (email: Li\_Zhengguo@a-star.edu.sg).  
}
\thanks{Chaobing Zheng and Wei Wang are with the school of Information Science and Engineering, Wuhan University of Science and Technology, Wuhan 430081, China. Chaobing Zheng is also with Advanced Remanufacturing and Technology Centre (ARTC), Agency for Science, Technology and Research (A*STAR)  (email: \{zhengchaobing,wangwei8\}@wust.edu.cn). }

}

\maketitle

\begin{abstract}
Block-wise varying exposures are effective for capturing appearance of a real-world high dynamic range (HDR) scene via a single-shot. However, brightness order reversal (BOR) artifacts, blocking artifacts and loss of details in highlight/shadow regions are  possible issues for block-wise varying exposures. In this paper, a novel neural block-wise auto-exposure algorithm is proposed to  address the issues. The proposed algorithm maps a temporary image and its exposure time to the exposure times of all non-overlapping  $16\times 16$ blocks. The temporary image can be either an sRGB or a raw image. Although the sRGB image is widely adopted in existing auto-exposure algorithms, exposure times of blocks in the highlight and shadow regions might not  be accurately learned due to the limited radiance measurement capability of the sRGB image.  This paper thus also explores the feasibility of  temporary image as a raw image. Due to the better radiance  measurement capability of the raw image, the exposure times of blocks are learned more accurately, and more details in the highlight/shadow regions of the real-world HDR scene are captured by the resultant HDR imaging system. Experiment results validate the proposed neural block-wise auto-exposure algorithm. 
\end{abstract}

\begin{IEEEkeywords}
High dynamic rang imaging, Single-shot, Block-wise, Auto-exposure, Neural network
\end{IEEEkeywords}

\section{Introduction}
Dynamic range is defined as the ratio of the maximum value to the minimum one that a certain quantity can assume. The dynamic range of a real-world high dynamic range (HDR) scene can reach 100,000,000:1 whereas the dynamic range of an 8-bit standard dynamic range (SDR) image is 256:1. Our human eyes can detect dynamic ranges varying from 10,000:1 to 1,000,000:1.  HDR imaging is thus highly demanded by many real-world applications \cite{1Somanath2021,1huang2022,1fran2016}. 

Merging differently exposed SDR images to expand the dynamic range is an effective way  for  the HDR imaging \cite{1debevec1997,Grossberg2003}. The differently exposed  SDR images are usually captured  sequentially by multiple shots. Unfortunately, ghosting artifacts caused by moving objects are an opening problem for the method, although there are many ghost removal algorithms \cite{1hu2012,11zheng2013,hu2013hdr,1li2014,oh2014robust,1kala2017,1wu2018,1yanq2020,1liu2021,1Prabhakar2021,1liuz2022,1zhangzl22024,chen2025ultrafusion}. Ghosting artifacts are not an issue for quasi-HDR algorithms on top of a single sRGB or raw image \cite{1kim2021,1endo2017,1liuyl2020,1zou2023}. Quasi-HDR algorithms directly hallucinate saturated details but the hallucinated details  are often implausible. Thus, they can only restore limited details in highlight/shadow regions of real-world HDR scene, and amplify noise in shadow region(s).

 New devices are emerging to capture an image which mimics the appearance of a real-world HDR scene without producing any ghosting artifacts by a single-shot \cite{1gu2010,1hirata2021,1tocci2011,1cannon20182,1xu2022}. Among them, a novel HDR imaging system with row-wise  exposures was proposed in \cite{1gu2010}. A temporary sRGB image is first captured for the HDR scene. An empirical equation between the temporary image and scene radiance is then built up. An optimal exposure is finally derived for each row on top of the empirical equation to minimize the number of over- and under-exposed pixels while keep most pixels well-exposed in the row. The first concern of the auto-exposure algorithm in \cite{1gu2010}  is that the temporary sRGB image has limited measurement capability of scene radiance. Rows in the highlight and shadow regions of the HDR scene could be overexposed and underexposed in the temporary image. It would be difficult or even impossible to accurately determine the exposure times of these rows to capture details of the real-world HDR scene. The second concern is that the exposure time of each row is determined independently. Brightness order reversal (BOR) and rowing artifacts could appear in the captured image. The third  concern is that the empirical equation might not accurately represent the nonlinear relationship between the temporary image and scene radiance. This affects the accuracy of the optimal exposure times. Fortunately, neural networks (NNs) are good at representing nonlinear functions.  Thus, it is  desired to develop a neural spatially-varying auto-exposure algorithm for single-shot HDR imaging.

In this paper, a novel neural block-wise auto-exposure algorithm is proposed such that  details especially those in highlight and shadow regions of a real-world HDR scene are captured with negligible BOR and blocking artifacts by a single-shot. The proposed algorithm is on top of a simple U-Net \cite{1ronne2015} and transformer with inputs as a temporary image and its exposure time and outputs as the exposure times of all non-overlapping $16\times 16$ blocks \cite{1hirata2021}.  The temporary image is captured to measure the radiance of each block for the real-world HDR scene. It can be  an sRGB  or a raw image. Although the sRGB image is widely adopted by existing auto-exposure algorithms \cite{1gu2010,1onzon2021,1tedla2023,1lee2024,1xu2025}, it is  challenging to learn the exposure times for $16\times 16$ blocks in highlight and shadow regions of  an HDR scene.  Fortunately, this problem can be addressed by the  raw image. Due to the better radiance measurement capability of the raw image, the exposure times are learned more accurately, especially for those blocks in the highlight and shadow regions. Subsequently, more details in the shadow and highlight regions are captured by the resultant imaging system. To the best of our knowledge, {\it we are the first to study the neural spatially varying auto-exposure with the temporary image as the raw image for single-shot HDR imaging}. 

A new data set is constructed from  the two existing data sets on HDR imaging in \cite{1xu2022} and \cite{1zheng2021A} as well as more sets of differently exposed SDR images captured by us. Camera response functions (CRFs) are first estimated for each set of differently exposed SDR images, and an HDR image is then synthesized  using the algorithm in \cite{1debevec1997}. The HDR image is a representation of the corresponding real-world scene.  Meanwhile, all 8-bit sRGB images in each set are fused together using an improved version of the multi-scale exposure fusion (MEF) algorithm in \cite{kou2017}. The fused image is selected as the target image to be captured by the proposed auto-exposure algorithm.   With the target image, the CRFs, and the HDR image, an optimal exposure time  will be derived for each pixel.  The optimal exposure times of all the pixels will serve as the ground-truth for the proposed neural auto-exposure algorithm.  Normally, large optimal exposure times dominate the learning process of the neural auto-exposure algorithm. It is thus difficult to approach those small optimal exposure times by the neural auto-exposure algorithm. Subsequently, the details in the highlight regions might not be well captured. A novel scale-invariant loss function is further proposed in this paper to address the problem. All optimal exposure times have the same importance regardless of their scales. As such, the proposed auto-exposure algorithm intends to approach all optimal exposure times regardless of their scales.  Subsequently, the captured sRGB image approaches the target image.  The BOR and blocking artifacts are also well addressed by the proposed algorithm.

The proposed block-wise auto-exposure algorithm is fundamentally different from the conventional auto-exposure algorithms in \cite{1onzon2021,1tedla2023,1lee2024,1xu2025} in the sense that the exposure time is the same for all blocks in an image by these algorithms. As such, they must determine whether highlight regions of a real-world HDR scene are overexposed or shadow regions are underexposed while this is not an issue for the proposed auto-exposure algorithm.  The proposed block-wise auto-exposure algorithm is also different from the row-wise auto-exposure algorithm in  \cite{1gu2010} in the sense that the algorithm in \cite{1gu2010} is a heuristic method with the temporary image as an sRGB image while ours is a learning-based method with the temporary image as either an sRGB or a raw image. 
Overall, three major contributions of this paper are:

 1) A new data-set is built up for studying the neural spatially varying auto-exposure which is important for HDR imaging. Temporary images and their exposure times, optimal exposure times for all the pixels, inverse CRFs, HDR images, fused 8-bit sRGB images are provided for other relevant HDR R\&D topics to use;
 
2) A simple neural block-wise auto-exposure algorithm and a novel scale-invariant loss function are proposed to determine an  exposure time for each non-overlapping $16\times 16$ block. Our supervised   exposure control is feasible as emerging specified circuits and systems such as dedicated digital signal processors \cite{1eki2021} for AI although it is the first attempt on the neural block-wise varying auto-exposure.

3) The neural block-wise auto-exposure with the temporary image as a raw image is a new initiative on HDR imaging. Due to the outstanding radiance measurement capability of the raw image, the exposure times of $16\times 16$ blocks in the shadow and highlight regions of HDR scenes are learned more accurately. More details in the shadow and highlight regions are captured.

The rest of this paper is organized as below. Related works are summarized in Section \ref{relevantworks}. A dataset is proposed in Section \ref{dataset} for studying spatially varying auto-exposure. Details on the proposed block-wise auto-exposure algorithm are given in Section \ref{neuralautoexposure}. Experiment results are provided in Section \ref{experimentresults} to validate the proposed auto-exposure algorithm. Finally, conclusion remarks are included in Section \ref{conclusionremarks}.

\section{Related Works}
\label{relevantworks}
A set of differently exposed SDR images can be captured to represent an HDR scene \cite{1debevec1997}. Existing HDR synthesis algorithms such as \cite{1debevec1997}  and MEF algorithms such as \cite{1mertens2007,kou2017,1ZGLi2017,MEFNet} assume that there are neither camera nor object movements in all input images. The assumption might not be true if the input images are captured sequentially by multiple shots \cite{1tao2026survey}. The fused image is blurred if there is camera movement and suffers from ghosting artifacts if there are moving objects. It is not difficult to align the input images with respect to one of them \cite{1akyuz2011,1wusq2014}
. However,  it is difficult or even impossible for existing ghost removal algorithms \cite{1hu2012,11zheng2013,hu2013hdr,1li2014,oh2014robust,1kala2017,1wu2018,1yanq2020,1liu2021,1Prabhakar2021,1liuz2022,1zhangzl22024,chen2025ultrafusion} to set up the correspondence among  pixels or features of two differently exposed images in the following scenarios: 1) background of the selected reference image includes large and heterogeneous over/under-exposed regions while the corresponding regions in other images are occluded by moving objects \cite{11zheng2013}; and 2) the reference image includes large and heterogeneous over/under-exposed regions with non-rigid movements.  Thus, ghosting artifacts are an open problem for multi-shot HDR imaging.
	
Many quasi-HDR imaging algorithms are proposed with input as a single sRGB or raw image \cite{1kim2021,1endo2017,1liuyl2020,1zou2023}. However, it is difficult for them to restore details in highlight and shadow regions of HDR scenes without amplifying noise in the shadow regions. Thus, it is highly demanded to develop single-shot HDR imaging \cite{1tao2026survey}. New devices were introduced to capture a real-world HDR scene by a single-shot \cite{1tocci2011,1gu2010,1cannon20182,1xu2022}. One example is a beam splitting based HDR video capturing system with a few sensors \cite{1tocci2011}.  Another one is a row-wise CMOS HDR video capturing system \cite{1gu2010}. An image is split into two fields with differently exposed times to simplify the CMOS sensor. The Canon released an innovative global shutter with a specific sensor that reads the sensor twice in an HDR mode. Every sensor pixel has two photodiodes: one small and one large. The small photodiode captures fewer photons and acts like a short exposure; the large one, a long exposure \cite{1cannon20182}. The differently exposed images are captured simultaneously using all these devices. Due to limited number of differently exposed images captured by the devices, the dynamic range of the captured image is limited. 

An effective solution is to divide an image into basic units which can be rows  \cite{1gu2010} or  $16\times 16$ blocks  \cite{1hirata2021}, and let the exposure time of every basic unit be adaptive to the corresponding scene radiance.   A temporary sRGB image is first captured to measure scene radiance of every row, and exposure time of every row is then determined by using a heuristic method \cite{1gu2010}. However, the accuracy of the heuristic method in \cite{1gu2010} is an issue. A possible solution is to determine the exposure times  using a simple NN which is good at representing a nonlinear function. This submission aims to explore this new solution by providing a neural spatially-varying auto-exposure algorithm for single-shot HDR imaging with the basic unit as a $16\times 16$ block. One more concern with the auto-exposure algorithm in \cite{1gu2010} is that the temporary image is an sRGB image which has limited radiance measurement capability. As such,  the exposure times of $16\times 16$ blocks in the highlight and shadow regions might not be accurately learned. This challenging problem is addressed in this paper using a raw image as the temporary image. Since the raw image has a  better radiance measurement capability than the  sRGB image, the exposure times of the blocks in the highlight and shadow regions would then be accurately learned.

\section{Problem Formulation on Block-wise Auto-exposure}
Each real-world HDR scene can be represented by a set of differently exposed SDR images \cite{1debevec1997}. Let $Z_i$ and $I_i$$(1\leq i\leq 2N-1)$ be the $i$th differently exposed sRGB and raw images of the HDR scene, respectively. 
Their exposure time is $\tau_{im}(i)$, and  $\tau_{im}(i)$ is monotonically increasing with respect to $i$. Neural auto-exposure with spatially varying exposures is required to capture the HDR scene by a single-shot, and it is a new R\&D problem for HDR imaging. 
 
 A temporary image is captured for the HDR scene to measure the radiance of all non-overlapping $16\times 16$ blocks and determine the desired exposure times for all blocks. For simplicity, the exposure time of the $l$th block is denoted as $\tau_{bl}(l)$. The temporary image can be either an sRGB or a raw image. For each set of differently exposed SDR images, the middle image $Z_N$ or $I_N$ is selected as the temporary image. {\it The inputs of the neural block-wise auto-exposure algorithm are the temporary image $Z_N$ or $I_N$ and its exposure time $\tau_{im}(N)$, and the outputs are $\tau_{bl}(l)$'s, the desired exposure times for all the non-overlapping $16\times 16$ blocks to maximize the dynamic range while minimize the BOR and blocking artifacts by the single-shot capture.} Mathematically, the mapping from the temporary image and its exposure time $\tau_{im}(N)$ to the optimal exposure times  $\tau_{bl}(l)$'s is learned as 
\begin{align}
\label{NNAutoExposure}
\{\tau_{bl}(l)\}=\Gamma_{\theta}(IM, \tau_{im}(N))\; ;\; IM\in \{Z_N, I_N\},
\end{align}
where $\Gamma_{\theta}(\cdot)$ is an NN with a small set of parameters $\theta$. 

The sRGB image is  widely adopted by existing auto-exposure algorithms \cite{1gu2010,1onzon2021,1tedla2023,1lee2024,1xu2025} as their temporary image, while the temporary image can be either the raw image $I_N$ or the sRGB image $Z_N$ in the new problem formulation.  The necessity of exploring the raw image as the temporary image is now elaborated. For example, consider two blocks $i$ and $j$ in  highlight regions. Both blocks are overexposed in the image $Z_N$. The radiance values of the pixels in block $i$ are much larger than those of the pixels in block $j$. The learned exposure time $\tau_{bl}(i)$ should be smaller than $\tau_{bl}(j)$. Unfortunately, these two blocks are not distinguished in the sRGB image $Z_N$.  Subsequently, the exposure times $\tau_{bl}(i)$ and $\tau_{bl}(j)$ cannot be accurately learned if the temporary image is $Z_N$. On the other hand, these two blocks could be distinguished in the raw image $I_N$, and the exposure times $\tau_{bl}(i)$ and $\tau_{bl}(j)$ could be  accurately learned if the temporary image is $I_N$. Two examples are given in Fig. \ref{myUNet}. 

\begin{figure*}[htb]
			\centering
			\includegraphics[width=0.95\textwidth]{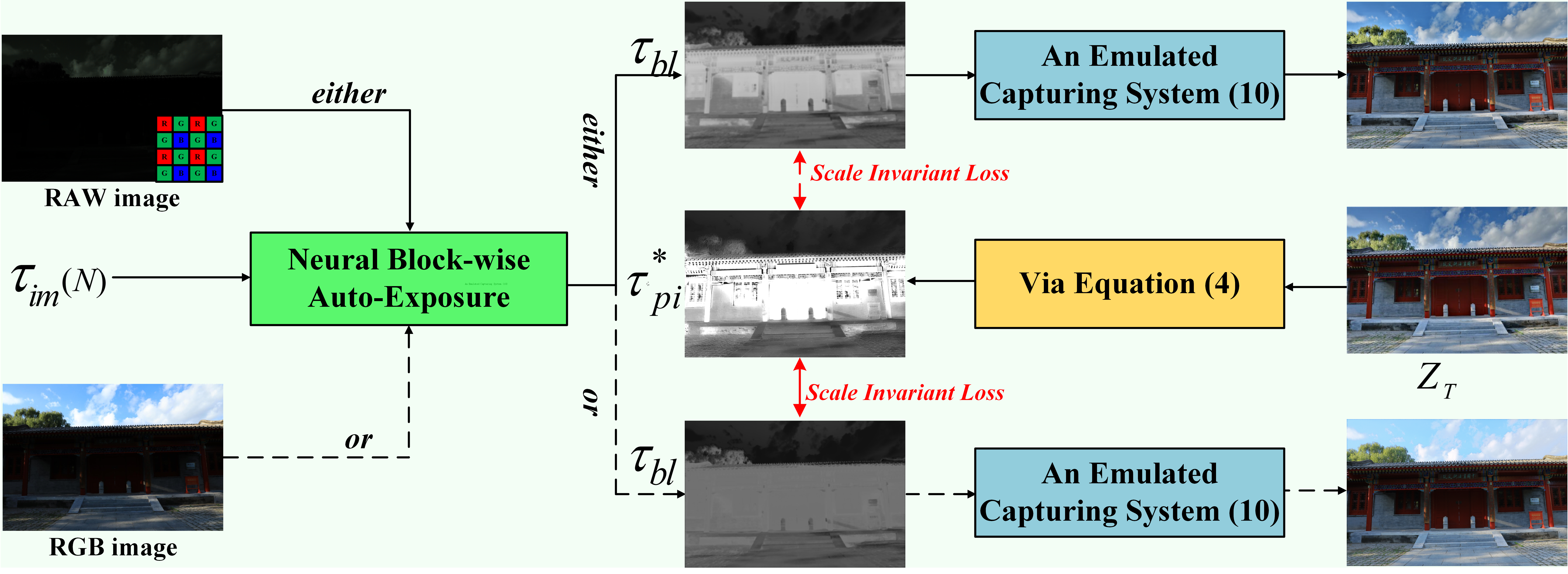}
			\caption{Diagram of the proposed neural block-wise auto-exposure scheme.  Inputs  are a temporary image which can be the sRGB image $Z_N$ or the raw image $I_N$ and its exposure time $\tau_{im}(N)$,  outputs are the exposure times of all non-overlapping $16\times 16$ blocks.  The ratios $\tau_{bl}(l)/\tau_{im}(N)$'s rather than the exposure times $\tau_{bl}(l)$'s are learned by the proposed algorithm. 
            }
    \label{framework}
 \end{figure*}

\section{Dataset for Neural Block-wise Auto-exposure}
\label{dataset}
Our proposed dataset is constructed from two existing datasets on HDR imaging in \cite{1xu2022,1zheng2021A} as well as more sets of differently exposed SDR images captured by us. The data set is constructed on the basis of the following observation:

An effective neural block-wise auto-exposure algorithm can be trained under the supervision of all pixels' optimal exposure times that are derived to minimize  blocking artifacts and  avoid BOR  artifacts caused by the auto-exposure algorithm. 

Similarly to the dateset in   \cite{1kala2017},  all sets of differently exposed images are captured from real-world scenes using different DSLR cameras. This is different from many existing data sets on HDR imaging which are synthesized. Each set in \cite{1xu2022} includes seven differently exposed SDR images while each set in \cite{1zheng2021A} has nine differently exposed SDR images. The exposure time of every image is the same for all pixels in the image. Both sRGB and raw images are available for each HDR scene. Their resolution is $2048\times 1024$.

Since the optimal exposure times of all pixels play an important role in the neural spatially-varying auto-exposure, details on the optimal exposure times are now presented. Let $E$ be the radiance of the HDR scene and $p$ be a pixel. Since the synthesized HDR image is the same for a set of differently exposed sRGB and raw images, it will be generated by using differently exposed sRGB images. The relationship between $Z_i(p)$ and $E(p)$ is represented by \cite{1debevec1997}
\begin{align}
Z_{i,c}(p)=f_c(E_c(p)\tau_{im}(i))\; ;\; c(\in\{R, G,B\}),
\end{align}
where $f_c(\cdot)$ is a CRF. 

With the $(2N-1)$ differently exposed SDR images $Z_i$'s and their exposure times $\tau_{im}(i)$'s, the inverse CRFs $f^{-1}_c(\cdot)$'s can be estimated using the algorithm in \cite{1debevec1997}. For simplicity, let $\ln(f^{-1}(\cdot))$ be denoted as $g(\cdot)$. The radiance value $E(p)$ can then be computed using the $Z_i(p)$'s, $\tau_{im}(i)$'s, and $g(\cdot)$ as
\begin{align}
\label{radiance}
\ln(E(p))=\frac{\sum_{i}w(Z_i(p))(g(Z_i(p))-\ln\tau_{im}(i))}{ \sum_iw(Z_i(p))},
\end{align}
where the function $w(z)$ is defined as
\begin{align}
w(z)=\left\{\begin{array}{ll}
z;&\mbox{if~}z\leq 128\\
255-z;&\mbox{otherwise}\\
\end{array}
\right..
\end{align}

The HDR image $E$ is a representation of the real-world HDR scene. All 8-bit sRGB images of each real-world HDR scene are fused by an improved version of the MEF algorithm in \cite{kou2017} and the resultant 8-bit sRGB image is denoted by $Z_T$. The image $Z_T$ is the target image to be approached by the proposed block-wise auto-exposure algorithm. 

For simplicity, $\tau_{pi}^*(p)$ denotes the optimal exposure time of the pixel $p$. It is computed as
\begin{align}
\nonumber
\tau_{pi}^*(p)=&\arg\min_{\tau}\{\|g(Z_T(p))-\ln(E(p)\tau)\|^2\}\\\label{deltatp*}
=&\exp(\frac{1}{3}\sum_c \log\frac{f^{-1}(Z_{T,c}(p))}{E_c(p)}).
\end{align}

The inverse CRFs $f^{-1}_c(\cdot)(c\in\{R, G, B\})$ can be implemented via look-up-tables (LUTs), and it is easy to compute $\tau_{pi}^*(p)$.  As shown in equation (\ref{deltatp*}), the optimal exposure time $\tau_{pi}^*(p)$ is determined by the radiance $E_c(p)(c\in \{R, G,B\})$. Thus, it is important to measure the radiance accurately. Unfortunately, it is very difficult or even impossible to do so via the heuristic method on top of the sRGB image in \cite{1gu2010}, especially for pixels in highlight and shadow regions of an HDR scene.

The proposed dataset is now built up. Each HDR scene contains: 1) the middle image $I_N$ (or $Z_N$) as well as its exposure time $\tau_{im}(N)$, 2) the optimal exposure times $\tau_{pi}^*(p)$'s, 3) the inverse CRFs $f^{-1}_c(\cdot)$'s, 4) the radiance map $E$,  and 5) the fused image $Z_T$. 
The dataset includes 420 HDR scenes, and is randomly split into three parts: 350 scenes for training, 30 scenes for verifying, and 40 scenes for testing. The size of our data set can be enlarged many times using cropping before training and testing. However, we choose random cropping during training to increase data diversity.

\section{The Proposed Neural Block-wise auto-exposure Algorithm}
\label{neuralautoexposure}

As shown in Fig. \ref{framework}, instead of learning the exposure times $\tau_{bl}(l)$'s directly, the ratios  $\tau_{bl}(l)/\tau_{im}(N)$'s are learned for all the non-overlapping  blocks by the proposed auto-exposure scheme. The objective is to increase the accuracy, convergence rate and robustness of the proposed algorithm.

\begin{figure}[htb]
\centering
\includegraphics[width=0.5\textwidth]{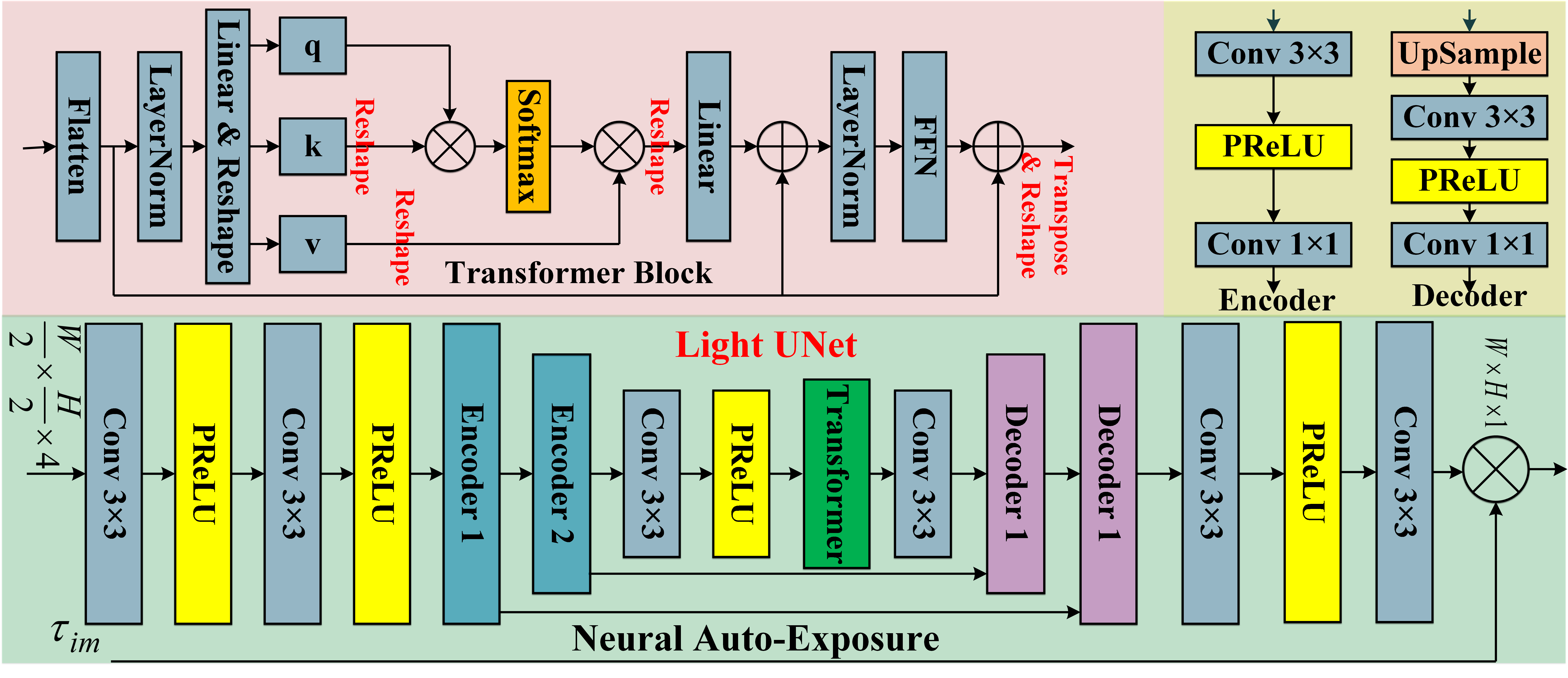}
    \caption{The proposed neural block-wise auto-exposure algorithm is on top of light-weighted U-Net and transformer which has about 0.6 M parameters. Inputs of the transformer are down-sampled channels. }
\label{UNet}
\end{figure}

The structure of the proposed neural network $\Gamma_{\theta}(\cdot)$ is shown in Fig. \ref{UNet}. The proposed algorithm is implemented using a simple U-Net \cite{1ronne2015} on top of $3\times 3$ convolution which is heavily used in modern CNN structures, a lite-transformer,  PReLU nonlinearities, downsampling via $2\times 2$ average pooling, and upsampling by $3\times 3$ deconvolution layers (transpose convolution). Since the speed is crucial for the neural auto-exposure algorithm \cite{1onzon2021}, the numbers of channels are a few in the proposed neural network $\Gamma_{\theta}(\cdot)$. Inputs of the transformer are down-sampled channels. 

\begin{figure*}[htb]
	\centering
	\includegraphics[width=1.0\textwidth]{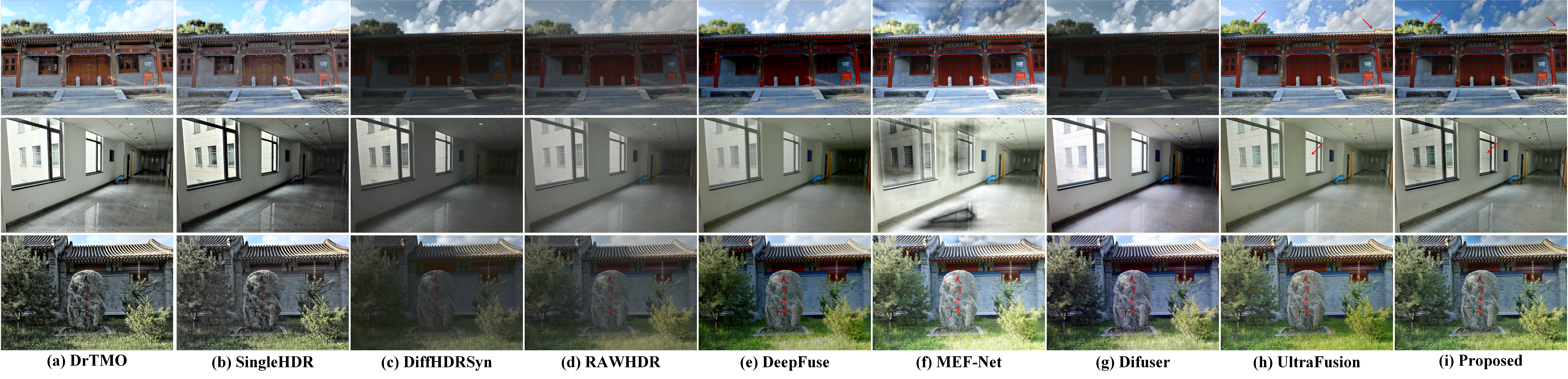}
	\caption{Comparison of the proposed algorithm with four quasi-HDR algorithms including DrTMO\cite{1endo2017}, SingleHDR \cite{1liuyl2020}, DiffHDRSyn \cite{1kim2021} and RAWHDR \cite{1zou2023} as well as four data-driven exposure fusion algorithms including DeepFuse \cite{1prabhakar2017}, MEF-Net \cite{MEFNet}, Difuser \cite{1tanglf2026} and UltraFusion \cite{chen2025ultrafusion} on three static HDR scenes. The proposed algorithm captures details in shadow and highlight regions without amplifying noise in shadow regions and producing halo and BOR artifacts.
    }
\label{morecomparison1}
\end{figure*}

\begin{figure*}[htb]
	\centering
	\includegraphics[width=1.0\textwidth]{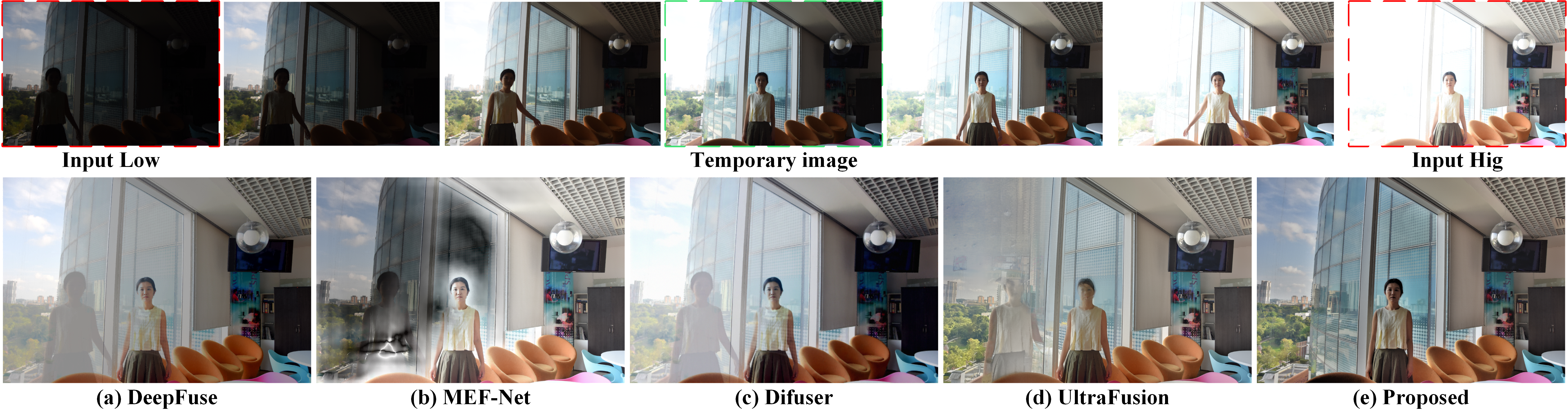}
	\caption{Comparison of the proposed algorithm with four data-driven exposure fusion algorithms including DeepFuse \cite{1prabhakar2017}, MEF-Net \cite{MEFNet},  Difuser \cite{1tanglf2026} and UltraFusion \cite{chen2025ultrafusion} on one dynamic HDR scene. The proposed algorithm captures details in shadow and highlight regions without  producing halo, BOR, and ghosting artifacts.}
\label{morecomparison2}
\end{figure*}

The loss function also plays an important role in the proposed algorithm.  Inspired by the well-known Weber fraction, a novel scale-invariant loss function $L(\tau_{bl})$ is proposed to train the proposed network (\ref{NNAutoExposure}), and the loss function is defined as 
{\small \begin{align}
\label{lossfunction1}
L=\frac{1+\alpha}{M}\sum_{l,p \in \Omega_l}(\ln\frac{\tau_{bl}(l)}{\tau^*_{pi}(p)})^2-\frac{1}{M^2}(\sum_{l,p \in \Omega_l}\ln\frac{\tau_{bl}(l)}{\tau^*_{pi}(p)})^2,
\end{align}}
where $\Omega_l$ is the $l$th non-overlapping $16\times 16$ block. $\tau_{bl}(l)$'s are the outputs of the proposed network  (\ref{NNAutoExposure}). $\tau_{pi}^*(p)$ is computed via equation (\ref{deltatp*}). $M$ is the total number of pixels in an image. $\alpha(<1)$ is a positive constant, and it is empirically selected as 0.2 in this paper.   

Let $|\Omega_l|$ be the cardinality of the set $\Omega_l$ and $\tau_{bl}^*(l)$ denote the optimal exposure time of $\Omega_l$, i.e. 
\begin{align}
\nonumber
\tau_{bl}^*(l)=&\arg\min_{\tau}\{\sum_{p \in \Omega_l}\|g(Z_T(p))-\ln(E(p)\tau)\|^2\}\\
\label{deltatl*}
=&\exp(\frac{1}{|\Omega_l|}\sum_{p\in \Omega_l}\ln\tau_{pi}^*(p)).
\end{align}

Let $M_b$ be the number of blocks in an image. 
 It can be derived that 
\begin{align}
\nonumber
 L=&\frac{1}{M_b}\sum_{l}(\ln\frac{\tau_{bl}(l)}{\tau_{bl}^*(l)})^2-\frac{1}{M_b^2}(\sum_{l}\ln\frac{\tau_{bl}(l)}{\tau_{bl}^*(l)})^2+\\\label{blocklevel}
&\frac{\alpha}{M_b}\sum_{l}(\ln\frac{\tau_{bl}(l)}{\tau_{bl}^*(l)})^2+\frac{1+\alpha}{M}\sum_{l,p\in \Omega_l}(\ln\frac{\tau_{pi}^*(p)}{\tau_{bl}^*(l)})^2.
\end{align}

Clearly, the proposed loss function (\ref{lossfunction1}) can be decomposed into three terms. The first term is to measure the variation among all $\ln \frac{\tau_{bl}(l)}{\tau_{bl}^*(l)}$' s. This term is minimized if and only if all $\frac{\tau_{bl}(l)}{\tau_{bl}^*(l)}$'s are the same. The second term is to measure the difference between $\ln\tau_{bl}(l)$ and $\ln\tau_{bl}^*(l)$ for all $l$'s.  The term is minimized if and only if $\tau_{bl}(l)$ is equal to $\tau_{bl}^*(l)$ for all $l$'s. In other words, {\it $\tau_{bl}(l)$ approaches $\tau^*_{bl}(l)$ regardless of the scale of $\tau^*_{bl}(l)$ for every block}. Equations (\ref{lossfunction1}) and (\ref{blocklevel}) imply that the loss function at the pixel level is the same as that at the block level. With the proposed loss function $L(\tau_{bi})$, all target exposure times $\tau_{pi}^*(p)$'s have the same importance regardless of their scales. This is very important for the proposed auto-exposure algorithm.

\begin{table*}[htbp]
	\setlength{\abovecaptionskip}{0pt}
	\setlength{\belowcaptionskip}{0pt}
	\centering
	\scriptsize
	\caption{Comparison of DrTMO \cite{1endo2017},  SingleHDR \cite{1liuyl2020}, and DiffHDRSyn \cite{1kim2021},  RAWHDR  \cite{1zou2023}, DeepFuse \cite{1prabhakar2017}, MEF-Net \cite{MEFNet}, Difuser \cite{1tanglf2026}, UltraFusion \cite{chen2025ultrafusion}  and the proposed 
    auto-exposure algorithm using MEF-SSIM, TMQI, HyperIQA and MUSIQ ($\uparrow$ larger is better) as well as NIQE ($\downarrow$ smaller is better).
		\label{tab3690}
	}
	\resizebox{2.0\columnwidth}{!}
	{
		\begin{tabular}{c|ccccccccc}
			\toprule[1pt]
			\textbf{Algorithm} & DrTMO  & SingleHDR  & DiffHDRSyn & RAWHDR & DeepFuse & MEF-Net & Difuser&  UltraFusion & Proposed \\
			\midrule
    \textbf{MEF-SSIM} $\uparrow$   & 0.8832       &0.8867    &0.7265   &0.5121    &0.8716    &0.8704   &0.7288   &{\it 0.9057}          & {\bf 0.9238}  \\\hline        
    \textbf{TMQI} $\uparrow$       & 0.8578      &0.8817    &0.7927   &0.7538    &0.8525    &0.8215  &0.8826   & {\it 0.9066}          & {\bf 0.9204}   \\\hline      
    \textbf{HyperIQA} $\uparrow$   & 68.03        &68.64     &60.34    &63.22     &68.79     &69.44     &65.02   &{\bf 70.89}           & {\it 70.45}    \\\hline   
    \textbf{MUSIQ} $\uparrow$      & 59.53         &60.50     &52.74    &51.31     &65.87     &66.30   &49.79    &{\bf 68.81}          & {\it 67.70}   \\\hline   
    \textbf{NIQE} $\downarrow$       & 2.9742       &3.0244    &3.4373   &2.8552    &2.6223    &2.5886   &3.9263   &{\it 2.5927}       & {\bf 2.5647}   \\\bottomrule[1pt]
		\end{tabular}
	}
\end{table*}

Our work is the first attempt on data-driven block-wise auto-exposure for single-shot HDR imaging. The proposed auto-exposure algorithm is different from the auto-exposure algorithm in \cite{1gu2010}. The algorithm in \cite{1gu2010} minimizes the number of over-exposed and under-exposed pixels within each row while keeping most pixels well-exposed in the row independently. It has a risk of producing BOR artifacts in a captured image. On the other hand, the optimal exposure times $\tau_{pi}^*(p)$ (or $\tau_{bl}^*$) serve as the ground truth for training the neural network $\Gamma_{\theta}(\cdot)$ for the proposed block-wise auto-exposure algorithm as shown in equation (\ref{lossfunction1}) (or (\ref{blocklevel})). The captured image is supposed to approach the targeted image $Z_T$. Therefore, {\it neither blocking nor BOR artifacts are an issue for the proposed neural block-wise auto-exposure algorithm}. The proposed  algorithm is also different from the capturing method in \cite{1xu2022} which only has two different exposure times for all rows in an image.

\section{Experiment Results}
\label{experimentresults}
In this section, the proposed auto-exposure algorithm with the temporary image as the raw image $I_N$ is compared with four state-of-the-art (SOTA) quasi-HDR algorithms, four SOTA data-driven exposure fusion algorithms, and two alternative block-wise auto-exposure algorithms. It should be mentioned that we implemented and tested the row-wise auto-exposure algorithm in \cite{1gu2010}.  Its performance is not good for the 40 testing scenes. Thus, the results are not included here.

\subsection{Comparison with Quasi-HDR and Exposure Fusion Algorithms}


The proposed auto-exposure algorithm  with the temporary image as the raw image $I_N$ is compared with four SOTA quasi-HDR algorithms and four SOTA data-driven exposure fusion algorithms. Among the four quasi-HDR algorithms, three are single-sRGB image based algorithms including  DrTMO \cite{1endo2017}, SingleHDR \cite{1liuyl2020}, and DiffHDRSyn \cite{1kim2021}, and one is a single-raw image based algorithm, i.e. RAWHDR  \cite{1zou2023}. The four exposure fusion algorithms are DeepFuse \cite{1prabhakar2017}, MEF-Net \cite{MEFNet}, Difuser \cite{1tanglf2026} and UltraFusion \cite{chen2025ultrafusion}. Inputs of the exposure fusion algorithms are two large-exposure-ratio (LER) images. Both static and dynamic HDR scenes are adopted to compare the algorithms.  

All algorithms are first compared subjectively. Readers are invited to view to the electronic version of the full-size figures and zoom in these figures in order to better appreciate the differences among images. As shown in Fig. \ref{morecomparison1},  the quasi-HDR algorithms such as DrTMO \cite{1endo2017} and SingleHDR \cite{1liuyl2020} suffer from noise amplification in the shadow region of the top scene. Although the data-driven exposure fusion algorithms avoid noise amplification in the shadow region, MEF-Net \cite{MEFNet} suffers from halo and BOR artifacts. Fine details in the shadow region of the top scene are not well preserved by DeepFuse \cite{1prabhakar2017} and Difuser \cite{1tanglf2026}.  The color of the fused images by UltraFusion \cite{chen2025ultrafusion} is slightly over-saturated for all scenes. Most of all, the data-driven exposure fusion algorithms produce ghosting artifacts for the dynamic scene as shown in Fig. \ref{morecomparison2}. This is because ghosting artifacts are an open problem for existing multi-shot HDR imaging solutions including UltraFusion \cite{chen2025ultrafusion}. All these problems are well addressed by the proposed auto-exposure algorithm. In addition,   the proposed auto-exposure algorithm preserves fine details in the highlight regions of the three HDR scenes in Figs \ref{morecomparison1} and \ref{morecomparison2} better than all quasi-HDR and exposure fusion algorithms. 

Two popular quality metrics in the field of HDR imaging: MEF-SSIM \cite{1ma2015} and  TMQI \cite{1yeganeh2013} are adopted to objectively compare all algorithms. The MEF-SSIM is computed using an output of each algorithm and all $Z_i$'s $(1\leq i\leq 2N-1)$ while the TMQI is computed using an output of each image and the HDR image which is synthesized from all $Z_i$'s $(1\leq i\leq 2N-1)$. As such, the MEF-SSIM and TMQI measure the similarity between an output of each algorithm and the real-world HDR scene.  NIEQ \cite{1mittal2013}, HyperIQA \cite{1Sus2020}, and MUSIQ \cite{1kej2021}  are also adopted to compare the algorithms. As shown in Table \ref{tab3690}, the proposed auto-exposure algorithm and UltraFusion \cite{chen2025ultrafusion} outperform  DrTMO \cite{1endo2017},  SingleHDR \cite{1liuyl2020}, and DiffHDRSyn \cite{1kim2021},  RAWHDR  \cite{1zou2023}, DeepFuse \cite{1prabhakar2017}, MEF-Net \cite{MEFNet},  and Difuser \cite{1tanglf2026} from the MEF-SSIM, TMQI, HyperIQA, MUSIQ and NIQE points of view.

Overall, the proposed auto-exposure algorithm outperforms DrTMO \cite{1endo2017}, SingleHDR \cite{1liuyl2020}, and DiffHDRSyn \cite{1kim2021}, RAWHDR  \cite{1zou2023}, DeepFuse \cite{1prabhakar2017}, MEF-Net \cite{MEFNet}, and Difuser \cite{1tanglf2026}  subjectively and objectively. The proposed auto-exposure algorithm captures fine details in the shadow/highlight regions of HDR scenes well without producing halo, BOR, and ghosting artifacts and amplifying noise in the shadow regions of HDR scenes.  

\begin{table*}[htbp]
\setlength{\abovecaptionskip}{0pt}
\setlength{\belowcaptionskip}{0pt}
\centering
\scriptsize
\caption{Comparison of different neural auto-exposure algorithms with temporary image as  $I_N$ using Abs Rel, Se Rel, Log MSE, SILong ($\downarrow$ smaller is better), MEF-SSIM \cite{1ma2015}, TMQI \cite{1yeganeh2013},  HyperIQA \cite{1Sus2020}, MUSIQ \cite{1kej2021} ($\uparrow$ larger is better), and NIEQ \cite{1mittal2013} ($\downarrow$ smaller is better).
\label{rawtab369}}
\resizebox{2\columnwidth}{!}
{
		\begin{tabular}{c|ccccccccc}
			\toprule[1pt]
			\textbf{Algorithm} &  Abs Rel $\downarrow$   &  Sq Rel $\downarrow$   & Log RMSE $\downarrow$   & SILog $\downarrow$  & MEF-SSIM $\uparrow$ & TMQI $\uparrow$ & HyperIQA $\uparrow$ & MUSIQ $\uparrow$ & NIQE $\downarrow$  \\
			\midrule
            1 (w transformer)       &   {\bf 0.1228}           & {\bf 0.0023}              & {\bf 0.1869}       &  {\bf 0.1825}        &  {\bf 0.9238}      &  {\bf 0.9204}    &  {\bf 70.45}     &  {\bf 67.70}      &  {\bf 2.5647} \\
            1 (w/o transformer)  &   0.1324           & 0.0026              & 0.1965       &  0.1927        &  0.9230      &  0.9186    &  70.18     &  67.90      &  2.5654 \\
		2                  &   1.6367           & 0.1202              & 1.0797       &  0.3525        &  0.8243      &  0.8048    &  67.26     &  62.97      &  3.9211\\
		3                  &   0.3674           & 0.0051              & 0.3822       &  0.3471        &  0.9198      &  0.9057    &  67.26     &  62.97      & 3.9211\\
			\bottomrule[1pt]
		\end{tabular}
	}
\end{table*}

\begin{figure*}[htb]
\centering	\includegraphics[width=1\textwidth]{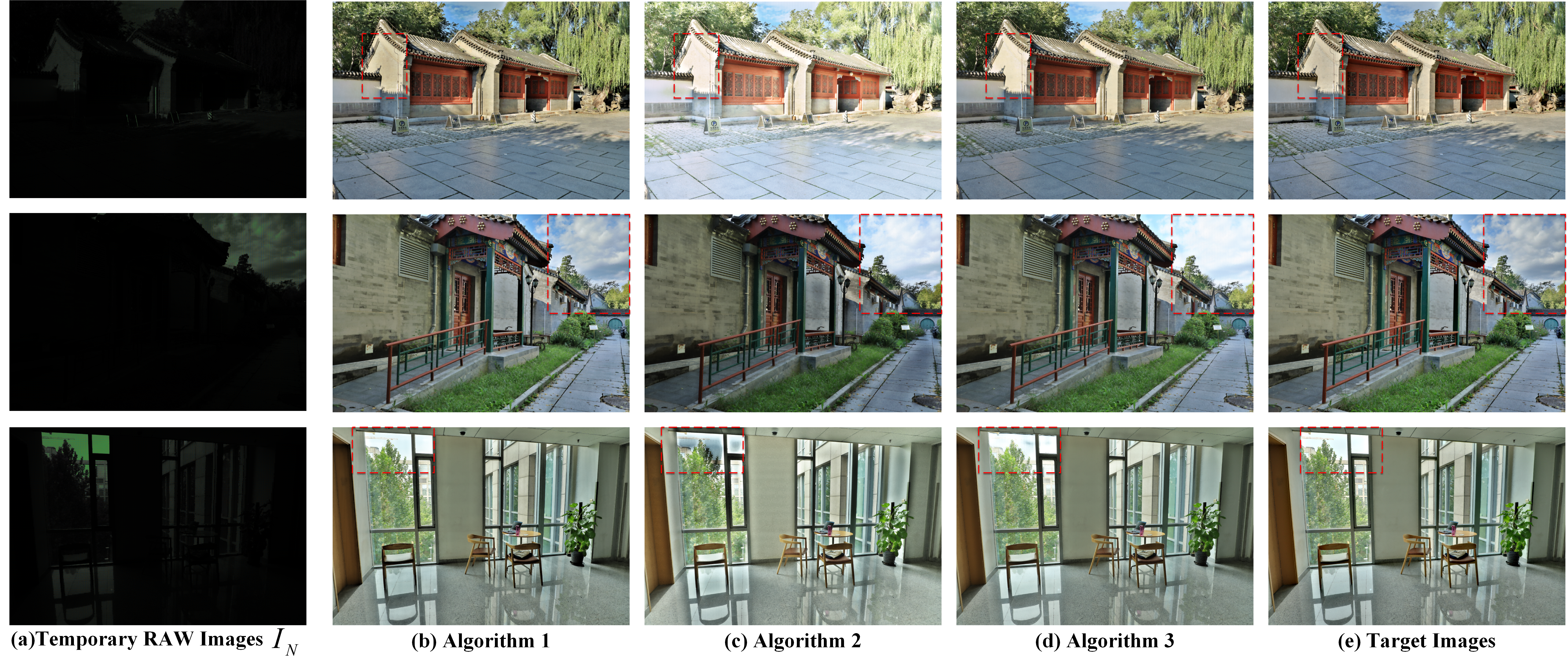}
	\caption{Subjective evaluation of three different auto-exposure algorithms with temporary image as raw image $I_N$. The captured images by Algorithm 1 approach the corresponding target images better than Algorithms 2 and 3.}
\label{lossfunction}
\vspace{-3mm}
\end{figure*}

\subsection{Comparison of Block-wise Auto-exposure Algorithms with Temporary Images as Raw Image $I_N$'s}

Since neural spatially-varying auto-exposure has not been widely studied for HDR imaging,  the following three different  auto-exposure algorithms are compared: 
{\it Algorithm 1}: learn $\frac{\tau_{bl}(l)}{\tau_{im}(N)}$'s via the loss function (\ref{lossfunction1}); {\it Algorithm 2}: learn $\tau_{bl}(l)$'s with the loss function (\ref{lossfunction1}); and {\it Algorithm 3}: learn $\frac{\tau_{bl}(l)}{\tau_{im}(N)}$'s with the loss function as 
\begin{align}
\label{lossfunction3}
\hat{L}(\tau_{bl})=\sum_{l,p\in \Omega_l}(\tau_{bl}(l)-\tau_{pi}^*(p))^2.
\end{align}

In this subsection, we consider the case that the temporary image is the raw image $I_N$. $\Gamma_{\theta}(I_N, \tau_{im}(N))$ in equation (\ref{NNAutoExposure}) is implemented as 
\begin{align}
\Gamma_{\theta}(I_N, \tau_{im}(N))=\tilde{\Gamma}_{\theta}(\frac{I_N}{\tau_{im}(N)}).
\end{align}

All these algorithms are trained using  the Adam optimizer with $\beta_1$ = 0.9 and $\beta_2$ = 0.99. The batch size is set at 8. The learning rate is initially set at $10^{-4}$ and then decreases using a cosine annealing schedule. All experiments are implemented using PyTorch on NVIDIA GP100 GPUs. These auto-exposure algorithms are first compared objectively using nine metrics: MEF-SSIM \cite{1ma2015}, TMQI \cite{1yeganeh2013},  HyperIQA \cite{1Sus2020}, MUSIQ \cite{1kej2021} as well as Abs Rel, Se Rel, Log MSE and SILong which are defined as 
\begin{align}
\mbox{Abs Rel}: & \frac{1}{M_b}\sum_{l}\frac{|\tau_{bl}(l)-\tau_{bl}^*(l)|}{\tau_{bl}^*(l)},\\
\mbox{Sq Rel}: & \frac{1}{M_b}\sum_{l}\frac{(\tau_{bl}(l)-\tau_{bl}^*(l))^2}{\tau_{bl}^*(l)}, \\
\mbox{Log RMSE}:  & \sqrt{\frac{1}{M_b}\sum_{l}(\ln\frac{\tau_{bl}(l)}{\tau^*_{bl}(l)})^2}, \\
\mbox{SILog}:  & \sqrt{\frac{1}{M_b}\sum_{l}(\ln\frac{\tau_{bl}(l)}{\tau^*_{bl}(l)}-\frac{1}{M_b}\sum_{l}\ln\frac{\tau_{bl}(l)}{\tau^*_{bl}(l)})^2}.
\end{align}

As shown in Table  \ref{rawtab369}, the first algorithm significantly outperforms the other two algorithms from the nine metrics points of view. 

In addition to the above objective evaluations, the three algorithms are also subjectively compared. The resultant exposure times by the three algorithms are adopted to capture an sRGB image that is emulated as follows:

With the HDR image $E$, and the CRFs $f_c(\cdot)(c\in\{R, G,B\})$, the captured image $\tilde{Z}$ can be generated as
\begin{align}
\label{capturingprocess}
\tilde{Z}_c(p)=f_c(E_c(p) \tau_{bl}(l))\; ;\; p\in \Omega_l,c\in\{R, G,B\}.
\end{align}
$\tilde{Z}_c(p)$ can be obtained as
$\arg\min_z\{|f^{-1}_c(z)-E_c(p) \tau_{bl}(l)|\}
$. To speed up the searching process, the initial value of $z$ can be set as $Z_{T,c}(p)$. The remaining searching process is carried out using the non-decreasing property of the inverse CRFs $f^{-1}(\cdot)$. 

\begin{table*}[htbp]
	\setlength{\abovecaptionskip}{0pt}
	\setlength{\belowcaptionskip}{0pt}
	\centering
	\scriptsize
	\caption{Comparison of different choices of $\alpha$ for the proposed auto-exposure algorithm with the temporary image as $I_N$ using Abs Rel, Se Rel, Log MSE, SILong ($\downarrow$ smaller is better), MEF-SSIM \cite{1ma2015}, TMQI \cite{1yeganeh2013},  HyperIQA \cite{1Sus2020}, MUSIQ \cite{1kej2021} ($\uparrow$ larger is better), and NIEQ \cite{1mittal2013} ($\downarrow$ smaller is better)
		\label{tab39}
	}
	\resizebox{2\columnwidth}{!}
	{
		\begin{tabular}{c|ccccccccc}
			\toprule[1pt]
			\textbf{$\alpha$}&  Abs Rel $\downarrow$   &  Sq Rel $\downarrow$   & Log RMSE $\downarrow$   & SILog $\downarrow$  & MEF-SSIM $\uparrow$ & TMQI $\uparrow$ & HyperIQA $\uparrow$ & MUSIQ $\uparrow$ & NIQE $\downarrow$  \\
			\midrule
             0                   &   0.1291           & 0.0025           & 0.1933       &  0.1886          &  0.7833     &  0.7563   &  61.63    &  57.01     &  2.6348\\
             0.2                 &   {\bf 0.1228}           &  {\bf 0.0023}          & {\bf 0.1869}       &  {\bf 0.1825}          &  {\bf 0.9238}     &  {\bf 0.9204}   &  {\bf 70.45}    &  {\bf 67.70}     &  {\bf 2.5647} \\
             1                   &   0.7822           & 0.0617           & 1.5570       &  0.1874          &  0.9229     &  0.9199   &  70.43    &  67.59     &  2.5692\\
							\bottomrule[1pt]
		\end{tabular}
	}
\end{table*}

\begin{table*}[htbp]
\setlength{\abovecaptionskip}{0pt}
\setlength{\belowcaptionskip}{0pt}
	\centering
	\scriptsize
	\caption{Comparison of the temporary image as $Z_N$ and $I_N$ using Abs Rel, Se Rel, Log MSE, SILong ($\downarrow$ smaller is better) as well as captured images using MEF-SSIM \cite{1ma2015}, TMQI \cite{1yeganeh2013},  HyperIQA \cite{1Sus2020}, MUSIQ \cite{1kej2021} ($\uparrow$ larger is better), and NIEQ \cite{1mittal2013} ($\downarrow$ smaller is better).
		\label{rawtab13369}
	}
    \resizebox{2\columnwidth}{!}{
	\begin{tabular}{c|ccccccccc}
		\toprule[1pt]
		\textbf{Temporary image} &  Abs Rel $\downarrow$   &  Sq Rel $\downarrow$   & Log RMSE $\downarrow$   & SILog $\downarrow$  & MEF-SSIM $\uparrow$ & TMQI $\uparrow$ & HyperIQA $\uparrow$ & MUSIQ $\uparrow$ & NIQE $\downarrow$  \\
		\midrule
		$Z_N$  &  0.1408  &  0.0029  &  0.2071  &  0.1976  &  0.9223  & 0.9130         &  70.37   &  67.57  &  2.5706\\
		$I_N$  &  {\bf 0.1228}  &  {\bf 0.0023}  &  {\bf 0.1869}  &  {\bf 0.1825}  &  {\bf 0.9238}  &  {\bf 0.9204}  &  {\bf 70.45}   &  {\bf 67.70}  &  {\bf 2.5647} \\
		\bottomrule[1pt]
	\end{tabular}
    }
\end{table*}

As demonstrated in Fig. \ref{lossfunction}, it is very difficult to learn $\tau_{bl}(l)$'s accurately by Algorithms 2 and 3. The captured image by Algorithm 3 could be brighter than the target image. This is not surprising. The optimal solutions to the loss functions (\ref{lossfunction1}) and (\ref{lossfunction3}) are 
$\tau_{bl}^*(l)$ in equation (\ref{deltatl*}) and $\frac{1}{|\Omega_l|}\sum_{p\in \Omega_l}\tau_{pi}^*(p)$, respectively. $\tau_{bl}^*(l)$ is usually smaller than $\frac{1}{|\Omega_l|}\sum_{p\in \Omega_l}\tau_{pi}^*(p)$. This problem is overcome by Algorithm 1 with the proposed loss function (\ref{lossfunction1}). Moreover, our capture system has neither visible blocking nor BOR artifacts. This is due to the optimal exposure times that are determined by the fused image $Z_T$. In general, Algorithm 1 captures more details in the shadow and highlight regions of HDR scenes. Therefore, Algorithm 1 outperforms the other two algorithms subjectively and objectively.

It should be pointed out that the emulated process is different from a real system because a raw image is first captured by the real one, and it is then converted to an sRGB image. The real system requires a new image signal processor (ISP). 

\subsection{Different Choices of Hyper-parameter $\alpha$}

$\alpha$ is an important hyper-parameter in the proposed loss function (\ref{lossfunction1}). Three different choices of $\alpha$ are compared in the end of this subsection.   

When $\alpha = 1.0$, the loss function is heavily weighted toward the absolute logarithmic MSE as shown in Eq. (\ref{lossfunction1}). Setting $\alpha = 0.2$ effectively suppresses the dominance of absolute scale errors, forcing the network to focus on optimizing relative exposure ratios in the logarithmic domain. When $\alpha = 0$, the loss function degrades into a purely scale-invariant variance metric. Although it maintains relative ratio consistency across blocks, it lacks constraints on the absolute scale, rendering the solution space under-constrained. As shown in Table \ref{tab39}, the choice of $\alpha$ as 0.2 outperforms the other two choices. Thus, $\alpha$ is selected as 0.2 in this paper.

\subsection{Evaluation of Auto-exposure Algorithms with Temporary Images as sRGB Image $Z_N$'s}
In the subsection, three different auto-exposure algorithms with the temporary images as $Z_N$'s are compared. $\Gamma_{\theta}(I_N, \tau_{im}(N))$ in equation (\ref{NNAutoExposure})  is implemented as 
\begin{align}
\label{NNAutoExposureRGB}
\Gamma_{\theta}(Z_N, \tau_{im}(N))=\tilde{\Gamma}_{\theta}(\frac{f^{-1}(Z_N)}{\tau_{im}(N)}).
\end{align}

\begin{figure}[htb]
\centering
\includegraphics[width=0.5\textwidth]{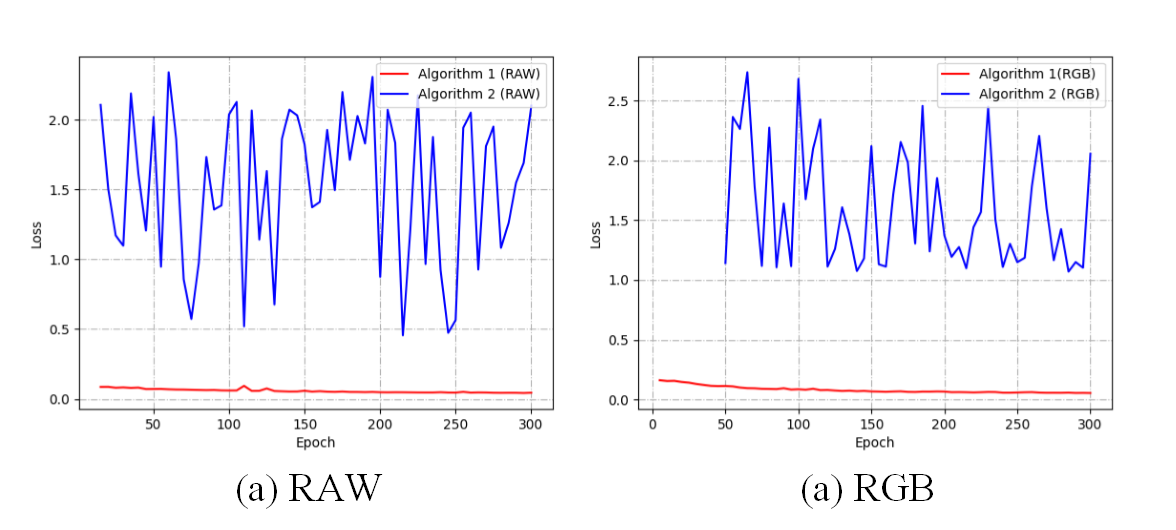}
    \vspace{-3mm}
    \caption{Training curves of three different algorithms with  temporary images as sRGB and raw images, respectively. It is difficult for Algorithm 2 to converge.}
\label{Loss}
\end{figure}

As demonstrated in Table \ref{tab369}, the accuracy of $\tau_{bl}(l)$'s is improved by learning the ratios $\tau_{bl}(l)/\tau_{im}(N)$'s than by learning the exposure times $\tau_{bl}(l)$'s directly. The proposed loss function (\ref{lossfunction1}) can further improve the accuracy of the exposure times $\tau_{bl}(l)$'s. The convergence rates of Algorithms 1 and 2 are also compared. As demonstrated in Fig. \ref{Loss}, Algorithm 1 converges faster than Algorithm 2 regardless of whether the temporary image is a raw image or an sRGB image. Thus, the proposed auto-exposure learns the ratios $\tau_{bl}(l)/\tau_{im}(N)$'s rather than the exposure times $\tau_{bl}(l)$'s.

\begin{table*}[htbp]
	\setlength{\abovecaptionskip}{0pt}
	\setlength{\belowcaptionskip}{0pt}
	\centering
	\scriptsize
	\caption{Comparison of different neural auto-exposure algorithms with the temporary image as  $Z_N$ using Abs Rel, Se Rel, Log MSE, SILong ($\downarrow$ smaller is better), MEF-SSIM \cite{1ma2015}, TMQI \cite{1yeganeh2013},  HyperIQA \cite{1Sus2020}, MUSIQ \cite{1kej2021} ($\uparrow$ larger is better), and NIEQ \cite{1mittal2013} ($\downarrow$ smaller is better).
		\label{tab369}
	}
	\resizebox{2\columnwidth}{!}
	{
		\begin{tabular}{c|ccccccccc}
			\toprule[1pt]
			\textbf{Algorithm} &  Abs Rel   &  Sq Rel  & Log RMSE  & SILog & MEF-SSIM & TMQI & HyperIQA & MUSIQ & NIQE  \\
			\midrule
              1    &  {\bf 0.1408}    &  {\bf 0.0029}     &  {\bf 0.2071}     &  {\bf 0.1976}   &  {\bf 0.9223}  & {\bf 0.9130}         &  {\bf 70.37}   &  {\bf 67.57}  &  {\bf 2.5706}\\
              2    &  3.5980    &  0.0806     &  1.5364     &  0.5653   &  0.8398  & 0.7730         &  64.99   &  61.33  &  3.1980\\
              3    &  0.2972    &  0.0066     &  0.3333     &  0.3156   &  0.9188  & 0.9075         &  70.11   &  67.56  &  2.5721\\
							\bottomrule[1pt]
		\end{tabular}
	}
\end{table*}

\begin{figure*}[htb]
\centering	\includegraphics[width=0.95\textwidth]{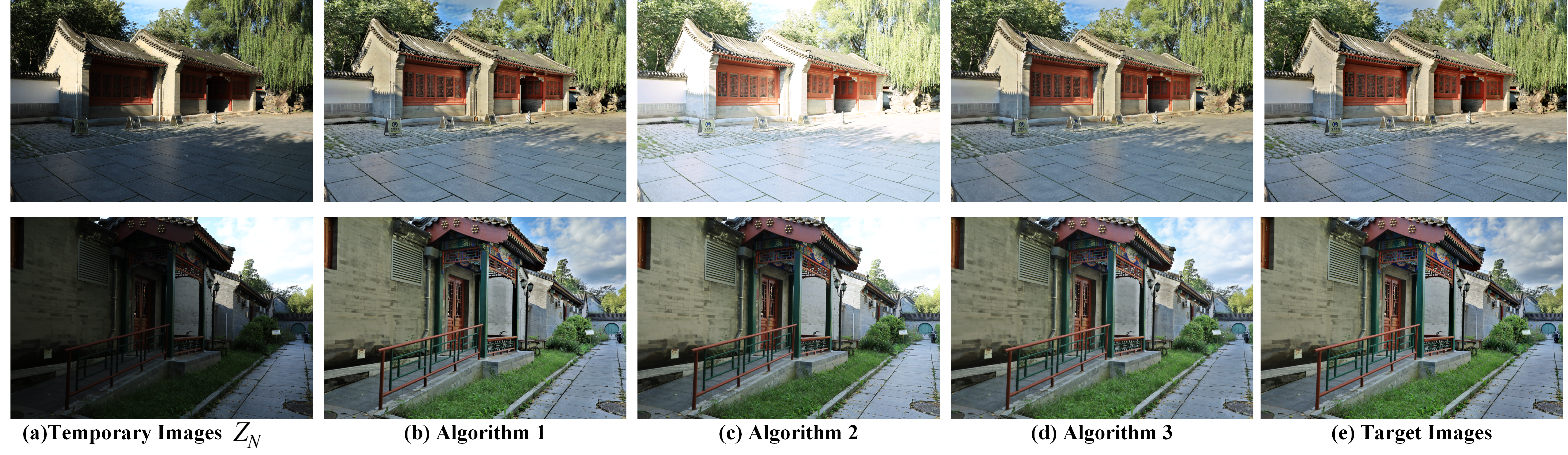}
	\caption{Subjective comparison of three auto-exposure algorithms with the temporary images as sRGB image $Z_N$. The captured images by Algorithm 1 approach the target images closer than the captured images by Algorithms 2 and 3.}
	\label{temporaryimageRGB}
\end{figure*}

These three algorithms are also subjectively compared. As shown in Fig.  \ref{temporaryimageRGB}, the captured images by Algorithm 3 are brighter than the captured images by Algorithm 1. Algorithm 1 outperforms Algorithms 2 and 3 in the sense that the captured images by Algorithm 1 approach closer to the target images than those captured images by Algorithms 2 and 3. 

\subsection{Comparison Between Two Types of Temporary Images}
\begin{figure}[htb]
\centering
\includegraphics[width=0.5\textwidth]{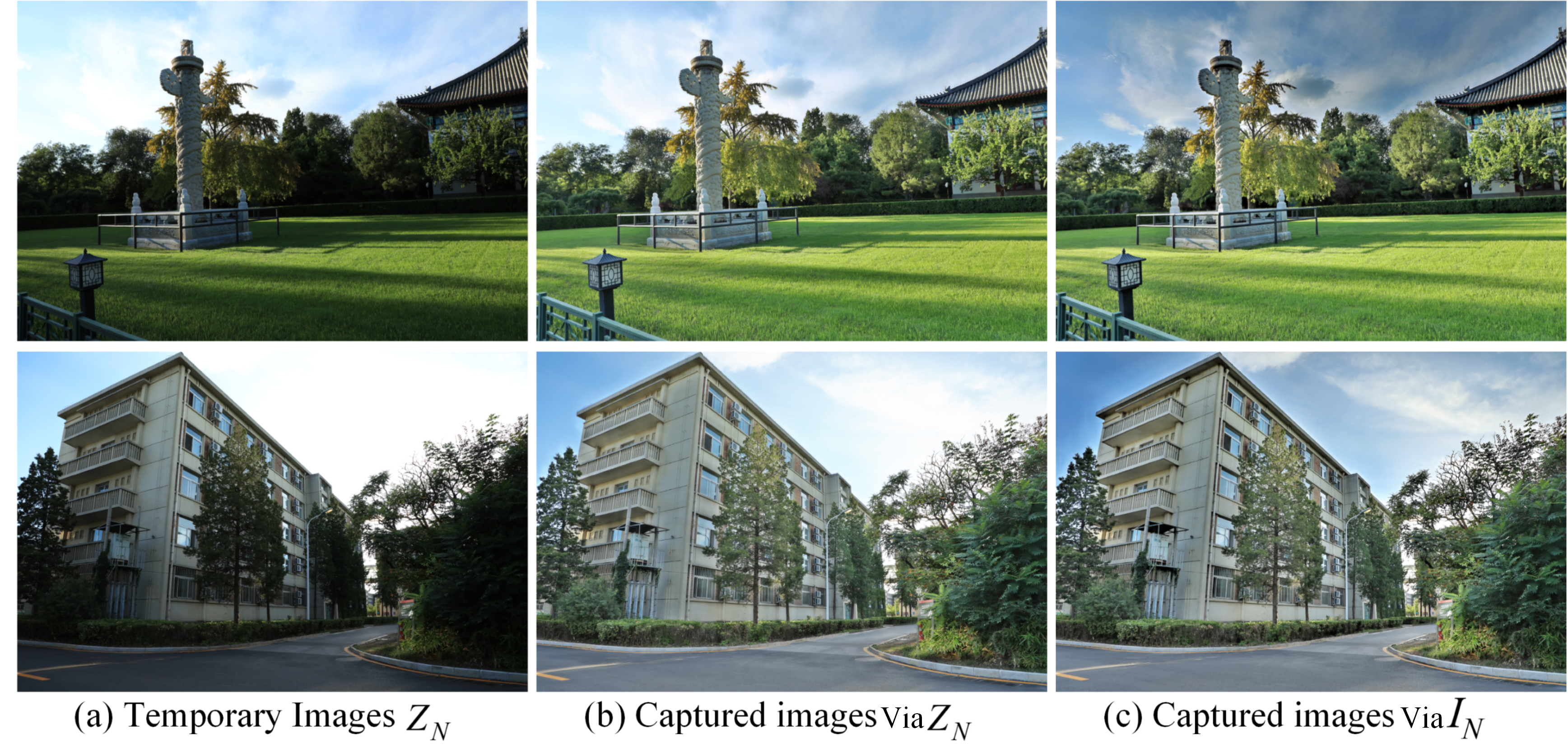}
    \vspace{-4mm}
    \caption{Subjective comparison of algorithm 1's with the temporary images as the raw image $I_N$ and the sRGB image $Z_N$, respectively. More details in the highlight regions are captured if the temporary image is the raw image $I_N$ due to the better radiance measurement capability of the raw image. }
\label{myUNet}
\end{figure}

Two different temporary images are compared in this subsection. As shown in  Table \ref{rawtab13369}, the raw image $I_N$ indeed outperforms the sRBG image $Z_N$ from all nine metrics points of view. In addition to comparing all the $16\times 16$ blocks, the experiment results for the $16\times 16$ blocks in the shadow and highlight regions are also compared for the sRGB image $Z_N$ and the raw image $I_N$ in the case where $\frac{\tau_{bl}(l)}{\tau_{im}(N)}$'s is learned via the loss function.  These blocks are identified by using their average values of the R, G, and B channels. If the average value is smaller than 6 or larger than 249, the block is included. As demonstrated in Table \ref{myrawtab1369}, the temporary image $I_N$ also outperforms the temporary image $Z_N$ from all four metrics points of view.  It is also shown Fig. \ref{myUNet} that more fine details in highlight regions are captured in both real-world HDR scenes with the temporary image as the raw image $I_N$. More comparison results are provided in our supplementary material.

\begin{table}[htbp]
\setlength{\abovecaptionskip}{0pt}
\setlength{\belowcaptionskip}{0pt}
\centering
\scriptsize
\caption{Comparison of blocks in the shadow and highlight regions between two choices of the temporary image as $Z_N$ and $I_N$.
		\label{myrawtab1369}
}
\resizebox{1\columnwidth}{!}
{
\begin{tabular}{c|cccc}
			\toprule[1pt]
			\textbf{Temporary image} & \textbf{ Abs Rel }  & \textbf{ Sq Rel } & \textbf{ Log RMSE }  & \textbf{SILog}  \\
			\midrule
            $Z_N$  &  0.4108    & 0.0138           & 0.4121      &  0.3574 \\
			$I_N$  &  {\bf 0.3813}    & {\bf 0.0098}           & {\bf 0.3832}      &  {\bf 0.3406}  \\
			\bottomrule[1pt]
		\end{tabular}
	}
\end{table}

\begin{figure}[htb]
\centering
\includegraphics[width=0.5\textwidth]{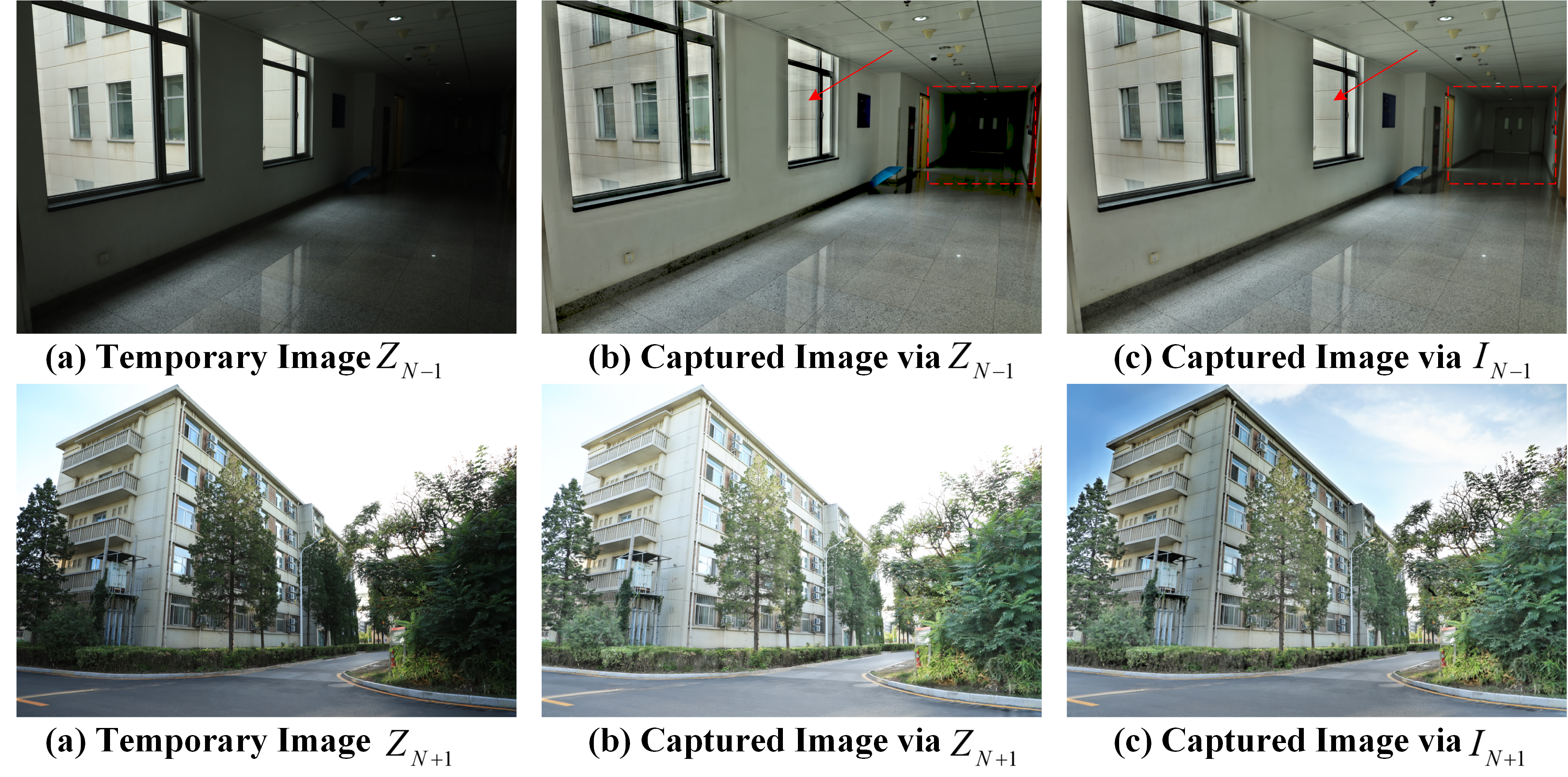}
    \vspace{-3mm}
    \caption{Subjecive evaluation of two HDR imaging systems with temporary image as an sRGB image and a raw image, respectively. The latter is much more robust than the former.}
\label{Robust}
\end{figure}

The robustness of two HDR imaging systems is also compared. Instead of using $Z_N$ (or $I_N$) as a temporary image, $Z_{N-1}$ (or $I_{N-1}$) and $Z_{N+1}$ (or $I_{N+1}$) are selected as temporary images. 
As shown in Fig. \ref{Robust}, the HDR imaging system with a raw image as the temporary image  is much more robust than  the HDR imaging system with an sRGB image as the temporary image. More details in the shadow/highlight regions are captured by the former. The latter also produces more blocking artifacts.

\subsection{Limitation of Our Auto-exposure Algorithm}

There is a gap between $\tau_{bl}(l)$'s and $\tau_{bl}^*(l)$ as demonstrated in Table \ref{tab369} and visible difference between the captured images and the target images as shown in Fig. \ref{temporaryimageRGB}. The gap can be reduced by the raw image $I_N$. Please find some elaboration as follows:

Since the speed and power consumption are crucial for auto-exposure of single shot HDR imaging \cite{1hirata2021}, the network $\Gamma_{\theta}(\cdot)$ is small. On the other hand, the network $\Gamma_{\theta}(\cdot)$ is supposed to: 1) restore the radiance $E$ from the middle image $I_N$ \cite{1zou2023} or $Z_N$; and 2) map the radiance $E$ to the optimal exposure time $\tau_{pi}^*$ (or $\tau_{bl}^*$) which is equivalent to convert the radiance E into the corresponding sRGB image  \cite{1vinker2021}. Both require deep neural networks as shown in \cite{1vinker2021,1zou2023}. Unfortunately,  deep neural networks are not affordable by real-time auto-exposure. Therefore, it is indeed very challenging to study neural block-wise  auto-exposure via the small network $\Gamma_{\theta}(\cdot)$.

\section{Conclusion Remarks}
\label{conclusionremarks}
Ghosting artifacts are an open problem for multi-shot high dynamic range (HDR) imaging. A novel neural block-wise auto-exposure algorithm has been introduced in this paper to capture an image which mimics the appearance of an HDR scene via a single-shot. Ghosting artifacts are not an issue for the new solution.  Since only a small neural network can be supported by real-time auto-exposure,  it is challenging to study the neural spatially-varying auto-exposure. Our work is the first attempt on the challenging problem. Better solutions would be expected to address this important problem. With the proposed auto-exposure algorithm, a raw image is captured for the HDR scene. An interesting R\&D problem is to design a neural ISP to covert the raw image into an sRGB image. This problem will be studied in our future research.

\bibliographystyle{IEEEtran}
\bibliography{main}

\end{document}